\documentclass[runningheads]{llncs}
\usepackage{graphicx}
\usepackage{xcolor}
\usepackage{multirow}
\usepackage{hhline}
\usepackage{subfig}
\usepackage{amsmath}

\usepackage{hyperref}

\hypersetup{
    colorlinks=true,
    linkcolor=blue,
    filecolor=magenta,      
    citecolor=blue,
    urlcolor=blue,
    }

\usepackage{bbold}

\usepackage{array}
\newcolumntype{P}[1]{>{\centering\arraybackslash}p{#1}}
\newcolumntype{M}[1]{>{\centering\arraybackslash}m{#1}}
\usepackage{amssymb}% http://ctan.org/pkg/amssymb
\usepackage{pifont}% http://ctan.org/pkg/pifont
\newcommand{\cmark}{\ding{51}}%
\begin{document}
\title{A Unified Model for Longitudinal Multi-Modal Multi-View Prediction with Missingness}

\titlerunning{UniLMMV}
% If the paper title is too long for the running head, you can set
% an abbreviated paper title here
%
\author{Boqi Chen, Junier Oliva, Marc Niethammer}
\authorrunning{B. Chen et al.}
% First names are abbreviated in the running head.
% If there are more than two authors, 'et al.' is used.
%
\institute{Department of Computer Science, University of North Carolina at Chapel Hill}
\maketitle              % typeset the header of the contribution
\begin{abstract}
Medical records often consist of different modalities, such as images, text, and tabular information. Integrating all modalities offers a holistic view of a patient's condition, while analyzing them longitudinally provides a better understanding of disease progression. However, real-world longitudinal medical records present challenges: 1) patients may lack some or all of the data for a specific timepoint, and 2) certain modalities or views might be absent for all patients during a particular period. 
In this work, we introduce a unified model for longitudinal multi-modal multi-view (MMMV) prediction with missingness. Our method allows as many timepoints as desired for input, and aims to leverage all available data, regardless of their availability. 
We conduct extensive experiments on the knee osteoarthritis dataset from the Osteoarthritis Initiative (OAI) for pain and Kellgren-Lawrence grade (KLG) prediction at a future timepoint. We demonstrate the effectiveness of our method by comparing results from our unified model to specific models that use the same modality and view combinations during training and evaluation. We also show the benefit of having extended temporal data and provide post-hoc analysis for a deeper understanding of each modality/view's importance for different tasks. Our code can be found at \href{https://github.com/uncbiag/UniLMMV}{https://github.com/uncbiag/UniLMMV}

% \keywords{Multi-modal \and Longitudinal \and Prediction}
\end{abstract}

\section{Introduction}
% Deep learning performs good but require large amount of data which can be hard in medical field.
In recent years, deep learning methods have revolutionized various domains, particularly in computer vision and natural language processing, owing to the accessibility of expansive datasets. The surge in available data has also catalyzed remarkable advances in medical data analysis~\cite{rotterdam,ukbiobank}, including image segmentation, registration, and prediction~\cite{chen2022recent}. 
However, analyzing medical records presents both opportunities and challenges. These records are often rich in modalities and views, spanning from demographic information to images of various regions and doctor's notes, that can provide multifaceted disease insights. However, they also pose challenges, ranging from the difficulty of acquiring all data to tracking patients over extended time periods. 

% Multi-modal is useful but often require modalities to be complete. Currently methods are hard to adapt for our case.
Multi-modal models have demonstrated remarkable efficacy using natural language, images, audio, etc.~\cite{imagebind,clip}. However, many existing approaches assume the simultaneous availability of all modalities during training and testing, which is not always realistic for medical records. Although strategies such as missing record synthesis~\cite{completer,liu2022assessing} have been explored to tackle this problem, few can handle large numbers of modalities or data captured from different patient regions. Moreover, these approaches require imputation, which is challenging by itself.

% Classification with longitudinal data provides extra benefit
%A notable characteristic of medical data is that longitudinal data might be available.
Longitudinal data often exists in medical applications, and the patient's historical trajectory holds invaluable clues for present diagnoses and future predictions. Many studies~\cite{cascarano2023machine,marti2020survey} use longitudinal data, but the issue of missingness, particularly in the context of longitudinal data and across multiple modalities or views, remains a substantial challenge.

% Overview of our model
In this work, we propose a novel and unified method to tackle the aforementioned challenges whilst bypassing the need for missing data imputation. 
The main contributions of our work are as follows: 
\begin{enumerate}
    \item We propose a unified model for longitudinal multi-modal multi-view prediction, offering flexibility in both the number of inputs and timepoints.
    \item We evaluate our approach on WOMAC pain~\cite{womac} and KLG~\cite{klg} prediction for osteoarthritis, where we elucidate the benefits of utilizing diverse modalities, views, and multiple timepoints.
    \item We demonstrate the generality of our unified model, which can handle different input combinations during evaluation.
    \item We conduct post-hoc analyses to assess the significance of each modality and view for different tasks.
\end{enumerate}

\section{Related Works}
% incomplete multi-modal classification
Medical records frequently encounter the issue of incompleteness, which can be categorized into missing at random (MAR), missing completely at random (MCAR), or missing not at random (MNAR). For the case of MAR, recent works have focused on synthesizing the missing data / features~\cite{completer,liu2022assessing} or learning joint multi-modal embeddings~\cite{dicnet,xu2022multi}, which then allow replacing one modality by the other. However, these methods are generally limited to data of the same subject or between a few modalities that can be effectively aligned. For diverse medical modalities and views, the most naive approach involves either removing samples with missingness or filling them in with special values. Recent works have also used a mask indicator to help ignore missing data~\cite{lmvcat,zhou2023incomplete}, offering a straightforward yet effective solution. In this work, we extend the masking-based strategies for complex missing patterns across longitudinal data.

% longitudinal data analysis
Longitudinal data analysis has been a popular area, especially in the medical field~\cite{cascarano2023machine,marti2020survey}. Historically, many studies~\cite{gibbons2010advances} have relied on traditional parametric statistical methods to analyze associations between variables, but they can have difficulty in capturing high-dimensional data. Recent advances in machine learning offer solutions to this challenge, with various innovative architectures being proposed. One of the simplest methods is through feature summarization~\cite{bhagwat2018modeling}, which aggregates all temporal information either at the input or in the feature space. More sophisticated techniques include applying recurrent models for disease prediction~\cite{cui2019rnn,hibehrt}. However, previous works on medical prediction often aim to produce a single output from a fixed number of inputs. In order to accommodate varying numbers of timepoints, we employ a transformer decoder model~\cite{gpt2}, enabling prediction at every timepoint, where each prediction depends only on the preceding inputs.

\section{Method}
\begin{figure}[t]
    \centering
    \includegraphics[width=.9\linewidth]{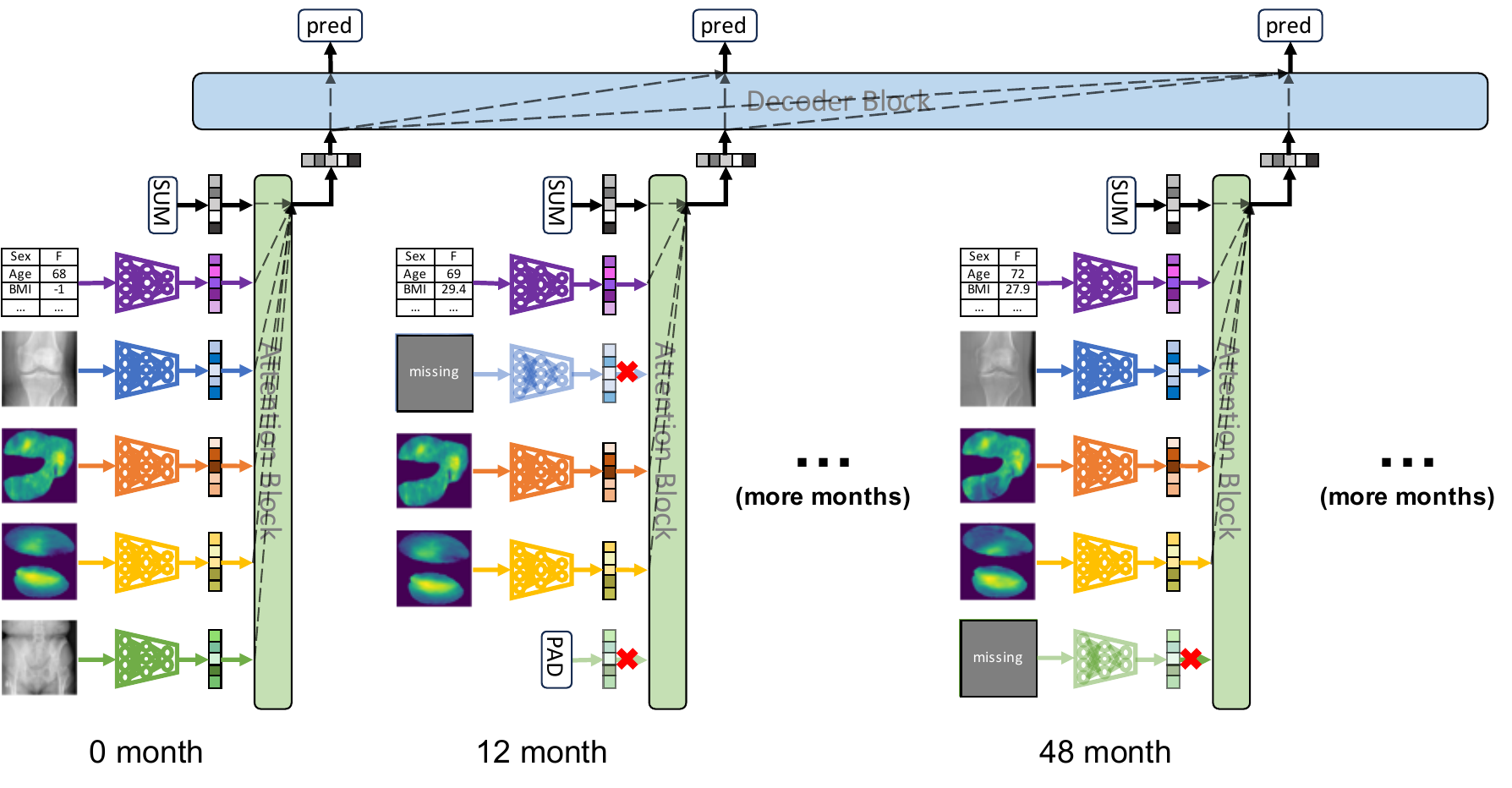}
    \caption{Our proposed model consists of an encoder for each modality and view, an attention block for summarizing the features, and a decoder block that predicts the result at each timepoint, focusing solely on previous data. [SUM] and [PAD] are learnable embeddings, where [SUM] outputs the summarized feature of all inputs, and [PAD] represents the modality or view that is absent for all patients.}
    \label{fig: Model}
\end{figure}

Our work presents a unified model for longitudinal MMMV prediction, as shown in Fig.~\ref{fig: Model}. 
Our novel model design is inspired by the following:
1) we want to process various types of modalities (e.g., tabular, radiography) and views (e.g., knee, pelvis); thus, we propose utilizing multiple encoders to represent information stemming from different modalities and views\footnote{For simplicity, we use view to represent both modality and view in the following.};
2) at any timepoint, there may be completely different patterns of missingness both within and between patients; thus, we propose utilizing a masked attention scheme to discard the missing data;
3) patients may have different numbers of available timepoints; thus, we propose a decoder model that can pay attention to various numbers of timepoints, each of which may consist of a subset of views.

\subsection{Feature Extraction}
We consider a dataset of multiple patients, which can be observed through multiple views at multiple timepoints.
Let $x_a[i,t]$ denote the $a$-th view ($a\in\{1, ..., n\}$) for the $i$-th patient at timepoint $t$. 
We use a neural network $\mathcal{F}_{\theta_a}$ to extract features from view $a$. We use convolutional neural networks for images and a transformer for tabular data, resulting in a feature vector:
\begin{equation}
    F_a[i,t]=\mathcal{F}_{\theta_a}(x_a[i,t])\,.
\end{equation}
We use $n$ different encoders for the $n$ different views. However, the same encoder is used for different timepoints of the same view.

\subsection{Feature Summarization}
To summarize features from various views, we use an attention block on the extracted features between all views $\{F_a[i,t]\}$. We first use a linear layer on the tabular feature vector to match the feature dimension of the image modalities. Subsequently, we ensure a uniform number of views by padding with a learnable [PAD] embedding. We also include a learnable [SUM] embedding, where its output serves as the summarized feature representing all views. The feature embeddings and their corresponding view embeddings (an embedding representing which view a feature belongs to) are added.
To support a subset of available views during evaluation, we randomly drop each view $50\%$ of the time during training of our unified model. For these dropped views, the mask indicator $\mathcal{M}$, which we introduce below, is set to $0$. We apply the attention block on all features (see Fig.~\ref{fig: Model}). The attention block includes multiple layers of transformer self-attention~\cite{transformer} blocks, where $Q, K, V$ represent the query, key, and value.

Given the possibility of missing views in medical records, we incorporate a mask indicator $\mathcal{M} \in \mathbb{R}^{n \times n} $ during training, where $\mathcal{M}_{i,j}=\{0,1\}$ represents absence or presence of the view. This allows us to manually assign a very low attention score $\delta=-1e^{9}$ to the missing view~\cite{lmvcat}, which ensures that the summarized feature does not focus on the missing data:

\begin{equation}
    \begin{split}
        Attention(Q, K, V) & = softmax\left(\frac{QK^T}{\sqrt{d_k}} \cdot \mathcal{M} + \delta \cdot (\textbf{1}- \mathcal{M})\right)V\,,
    \end{split}
\end{equation}
where $d_k$ is the dimension of the key embedding, $Q=W_Q X$, $K=W_K X$, $V=W_V X$ with $X$ the input embedding features and $\{W_Q,W_K,W_V\}$ learnable parameters.  The final summarized feature $F_{[SUM]}[i, t]$ is obtained after multiple layers, each with a combination of multiple heads of the above formula.

\subsection{Longitudinally-Aware Prediction}
Following the extraction of a summarized feature at each timepoint, our transformer decoder block disregards future timepoints~\cite{gpt2}, and the prediction is based solely on preceding timepoints:

\begin{equation}
    p(o_{i,t}) = p(o_{i,t}|F_{[SUM]}[i, \leq t])\,,
\end{equation}

where $o_{i,t}$ is the output at timepoint $t$  and $F_{[SUM]}[i, \leq t]$ is the set of all summarized features of patient $i$ with timepoint $\leq t$. In scenarios where labels of certain timepoints are missing, we do not consider these predictions when calculating the loss. 
We use a weighted cross-entropy loss that solely considers instances with available prediction labels:

\begin{equation}
    loss = 
\begin{cases}
    - \frac{1}{m \cdot l - |\mathcal{D}|}  {\sum_{i=1}^{m} \sum_{t=1}^{l}  \mathbb{1}_{(i,t) \notin \mathcal{D}} \cdot w_{i,t} \cdot y_{i,t} \cdot log(p(o_{i,t})) }\,,& \text{if } m \cdot l > |\mathcal{D}|\\
    0,              & \text{otherwise}\,,
\end{cases}
\end{equation}

where $m$ is the number of elements in a mini-batch, $l$ is the total number of timepoints, $\mathcal{D}$ is the set of the ignored samples in the mini-batch, $w$ is the weight for balancing each class, and $y$ is the given label.

\section{Experimental Results}

\subsection{Dataset}
We evaluate our approach using the OAI dataset\footnote{https://nda.nih.gov/oai/}, which contains $4,796$ patients between $45$ to $79$ years old at the time of recruitment. Each patient is longitudinally followed for up to $96$ months, with separate evaluations for the left and right knees. Tabular data exists for all patients at every timepoint, but may contain missing attributes. Image data are less complete. See Appendix~\ref{sec: OAIOverview} for the distribution of input image data at different timepoints. 

Our goal is to predict outcomes $24$ month ahead, thus our inputs only include data up to $72$ month. We use $6$ timepoints for pain prediction and $5$ timepoints for KLG prediction due to the absence of KLG labels at the $60$ month timepoint. We randomly select $50\%$ of the patients for training, $12.5\%$ for validation, and the remainder for testing. We conduct all experiments $5$ times with different seeds for training and report the mean $\pm$ standard deviation for average precision (AP), AUC ROC (ROC)\footnote{AP and ROC are originally defined for binary classes. For KLG prediction with $4$ classes, we use the one-vs-rest scheme.}, and macro accuracy (Macro ACC) for combinations between tabular (T), femoral and tibial cartilage thickness maps(C), knee radiography (K), and pelvis radiography (P).

We evaluate our method on WOMAC pain and KLG prediction, where the labels range from $0\sim20$ and $0\sim4$, respectively. For pain prediction, we define WOMAC$<5$ as no pain and the rest as pain, while for KLG prediction, we merge KLG$=0 \& 1$ since osteoarthritis is considered definitive only when KLG$\geq2$~\cite{kohn2016classifications}.

\begin{figure}[t]
    \centering
    \includegraphics[width=.95\linewidth]{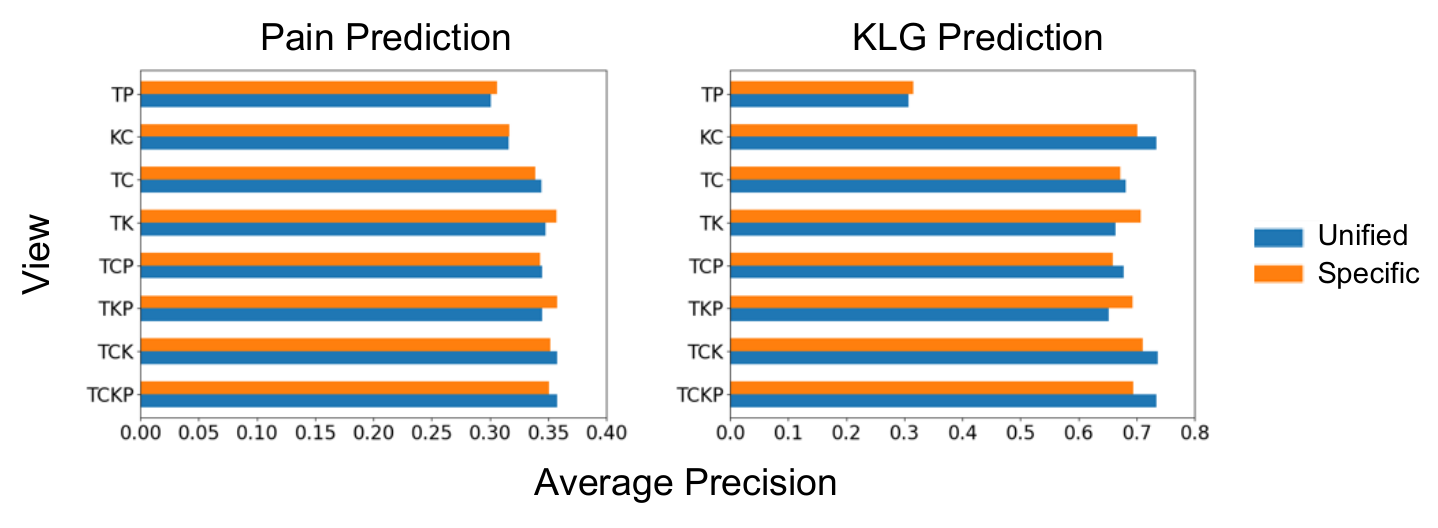}
    \caption{Comparison of average precision scores between view-specific models and the models obtained via modality dropout from our unified model. The y-axis represents the combination of different views, e.g., TCKP represents using {\bf t}abular, {\bf c}artilage thickness maps, {\bf k}nee radiography, and {\bf p}elvis radiography.}
    \label{fig: ModalityRemoval}
\end{figure}

\subsection{Data Preprocessing}
Our dataset includes both tabular data and images. All images are resized to $128 \times 128$ for feature extraction with augmentation of randomly rotating up to 15 degrees and adding Gaussian noise. Data preprocessing details for each view are described below.

\noindent\textbf{Tabular (T).}
We filter the tabular attributes from ~\cite{keefe2023patterns} and keep only those that can be easily captured, leaving us with $17$ attributes as detailed in Appendix~\ref{sec: Tabular}. An additional `side' indicator is added to indicate if the left or the right knee is used for prediction. We fill in the missing entries with $-1$ for continuous attributes and `Missing' for categorical attributes. Subsequently, all categorical values are encoded into numerical values for the ease of feature extraction.

\noindent\textbf{Cartilage Thickness Maps (C).}
The femoral and tibial cartilage thickness maps are not directly available in the OAI dataset. They are extracted from the DESS MR images through cartilage segmentation, mesh extraction, registration to a common atlas, and 2D thickness projection as detailed in~\cite{dadp}.

\noindent\textbf{Knee Radiography (K).}
The knee radiographs encompass substantial areas of the femur and tibia. To extract the joint region, we employ the method proposed in~\cite{kneel} for keypoint extraction. Then, a region of $140mm \times 140mm$ is extracted. To mitigate potential noise introduced during image acquisition, we normalize the radiographs by linearly scaling the intensities such that the smallest $99\%$ of values map to $[0, 0.99]$. Additionally, we apply horizontal flipping to all right knees~\cite{mris} and add random contrast adjustments.

\noindent\textbf{Pelvis Radiography (P).}
For pelvis radiographs, we extract a region of $350mm \times 400mm$ through center cropping. Similar to the knee radiographs, we apply normalization and contrast adjustments during augmentation. We do not flip images because the pelvis radiographs show the entire pelvis, including both the left and right sides. 

\subsection{Network Training}
We use the SAINT model~\cite{saint} to extract features from tabular data and the ResNet18 model~\cite{resnet} for image feature extraction.
During training, we freeze the first ResNet18 block to avoid overfitting but train all other blocks. We initialize ResNet18 using the pretrained  ImageNet~\cite{imagenet} parameters and train the SAINT model from scratch. Both our attention and decoder blocks use $6$ layers of self-attention, each containing $8$ heads. 
For all experiments, we train for $30$ epochs with a batch size of $256$. We use AdamW~\cite{adamw} as the optimizer and the one cycle scheduler~\cite{onecycle} with a maximum learning rate of $1e-6$ for the tabular and cartilage thickness maps encoders and $1e-5$ for all remaining components for pain prediction. We increase the learning rate of all image encoders by a factor of $10$ for KLG prediction. The network parameters resulting in the best average precision scores on the validation set were selected for testing.

\begin{table}[t]
    \begin{center}
    \resizebox{\textwidth}{!}{
    \begin{tabular}{|M{0.3cm}|M{0.3cm}|M{0.3cm}|M{0.3cm}|M{2.0cm}|M{2.0cm}|M{2.0cm}|M{2.0cm}|M{2.0cm}|M{2.0cm}|}
        \hline
        \multicolumn{4}{|c|}{Views} & \multicolumn{3}{c|}{Pain Prediction} & \multicolumn{3}{c|}{KLG Prediction} \\
        \hline
        T & C & K & P & $72$ & $24 \rightarrow 72$ & $0 \rightarrow 72$ & $72$ & $24 \rightarrow 72$ & $0 \rightarrow 72$ \\
        \hline
        \cmark & & & & $0.282 \pm 0.005$ & $0.309 \pm 0.005$ & $\textbf{0.321} \pm \textbf{0.006}$ & $\textbf{0.269} \pm \textbf{0.002}$ & $0.268 \pm 0.003$ & $0.269 \pm 0.002$ \\
        \hline
        & \cmark & & & $0.270 \pm 0.005$ & $0.276 \pm 0.003$ & $\textbf{0.280} \pm \textbf{0.002}$ & $0.513 \pm 0.011$ & $0.526 \pm 0.007$ & $\textbf{0.542} \pm \textbf{0.006}$ \\
        \hline
        & & \cmark & & $0.289 \pm 0.002$ & $0.304 \pm 0.004$ & $\textbf{0.307} \pm \textbf{0.003}$ & $0.525 \pm 0.014$ & $0.549 \pm 0.014$ & $\textbf{0.555} \pm \textbf{0.014}$ \\
        \hline
        \cmark & \cmark & & & $0.314 \pm 0.006$ & $0.329 \pm 0.006$ & $\textbf{0.337} \pm \textbf{0.006}$ & $0.532 \pm 0.019$ & $\textbf{0.557} \pm \textbf{0.012}$ & $0.556 \pm 0.014$ \\
        \hline
        \cmark & & \cmark & & $0.319 \pm 0.004$ & $0.340 \pm 0.003$ & $\textbf{0.348} \pm \textbf{0.004}$ & $0.531 \pm 0.004$ & $0.553 \pm 0.011$ & $\textbf{0.555} \pm \textbf{0.015}$ \\
        \hline
        \cmark & & & \cmark & $0.280\pm0.007$ & $0.311\pm0.005$ & $\textbf{0.321} \pm \textbf{0.006}$ & $0.269\pm0.003$ & $0.276\pm0.004$ & $\textbf{0.282} \pm \textbf{0.005}$ \\
        \hline
        & \cmark & \cmark & & $0.291 \pm 0.006$ & $0.298 \pm 0.003$ & $\textbf{0.303} \pm \textbf{0.002}$ & $0.528 \pm 0.018$ & $0.550 \pm 0.021$ & $\textbf{0.559} \pm \textbf{0.013}$ \\
        \hline
        \cmark & \cmark & \cmark & & $0.325 \pm 0.004$ & $0.337 \pm 0.003$ & $\textbf{0.342} \pm \textbf{0.003}$ & $0.559 \pm 0.017$ & $0.580 \pm 0.013$ & $\textbf{0.587} \pm \textbf{0.015}$ \\
        \hline
        \cmark & \cmark & & \cmark & $0.312\pm0.006$ & $0.337\pm0.005$ & $\textbf{0.345} \pm \textbf{0.005}$ & $0.532\pm0.013$ & $0.567\pm0.007$ & $\textbf{0.576} \pm \textbf{0.016}$ \\
        \hline
        \cmark & & \cmark & \cmark & $0.318\pm0.013$ & $0.342\pm0.005$ & $\textbf{0.349} \pm \textbf{0.004}$ & $0.526\pm0.013$ & $0.548\pm0.021$ & $\textbf{0.550} \pm \textbf{0.019}$ \\
        \hline
        & \cmark & \cmark & \cmark & $0.290\pm0.002$ & $0.300\pm0.002$ & $\textbf{0.305} \pm \textbf{0.004}$ & $0.539\pm0.021$ & $0.550\pm0.026$ & $\textbf{0.567} \pm \textbf{0.021}$ \\
        \hline
        \cmark & \cmark & \cmark & \cmark & $0.322\pm0.002$ & $0.333\pm0.007$ & $\textbf{0.339} \pm \textbf{0.007}$ & $0.541\pm0.020$ & $0.573\pm0.013$ & $\textbf{0.579} \pm \textbf{0.015}$ \\
        \hhline{|=|=|=|=|=|=|=|=|=|=|}
        \cmark & \cmark & \cmark & \cmark & $0.327 \pm 0.007$ & $0.342 \pm 0.007$ & $\textbf{0.346} \pm \textbf{0.003}$ & $0.586 \pm 0.015$ & $\textbf{0.606} \pm \textbf{0.014}$ & $0.600 \pm 0.012$ \\
        \hline
    \end{tabular}
    }
    \end{center}
    \caption{Mean $\pm$ STD of the average precision score for pain and KLG prediction for $96$ month given varying numbers of previous timepoints. $X \rightarrow Y$ represents using timepoints starting from month $X$ to month $Y$. The last row presents results from our unified model, and all others are from our view-specific model.}
    \label{tab: Pred96m}
\end{table}

\subsection{Results}
In this section, we show our experimental results by investigating the following:

\noindent\textit{Can our unified model, trained with all available views, perform on par with view-specific models during evaluation?}
To assess the generalizability of our unified model to fewer views during testing, we excluded up to two views at a time and compared the results with view-specific models (results for each model can be found in Appendix~\ref{sec: PredResult}). 
Fig.~\ref{fig: ModalityRemoval} and Appendix~\ref{sec:AdditionalPruning} show bar chart comparisons. We observe that our unified model performs on par with view-specific models while providing the flexibility to use views whenever available. %This can be attributed to the benefit of including additional views during training, which provides a better feature extraction.
Note that when cartilage thickness maps are not available, our unified model performs slightly worse than view-specific models but generally better when they are present (regardless of what other views are missing). In particular, this is the case for KLG prediction. We hypothesize that this is due to the ease in predicting higher KLG (which involves cartilage degradation/thinning) from cartilage thickness maps. Further insights into this phenomenon are provided below.

\noindent\textit{Does an extended observation period enhance prediction results?}
In order to show the benefit of using longitudinal data, we predict results at the final $96$ month timepoint, varying the number of previous timepoints provided. Tab.~\ref{tab: Pred96m} shows that in most cases, it is beneficial to include an increased number of timepoints, underlining the importance of longitudinal data in improving predictive accuracy. However, few exceptions appear for predicting KLG with the inclusion of tabular data. This can be attributed to the limitations of tabular data in providing information for KLG, which is scored only based on knee radiographs.

\begin{figure}[!tbp]
  \centering
  \subfloat[]{
      \includegraphics[width=0.32\textwidth]{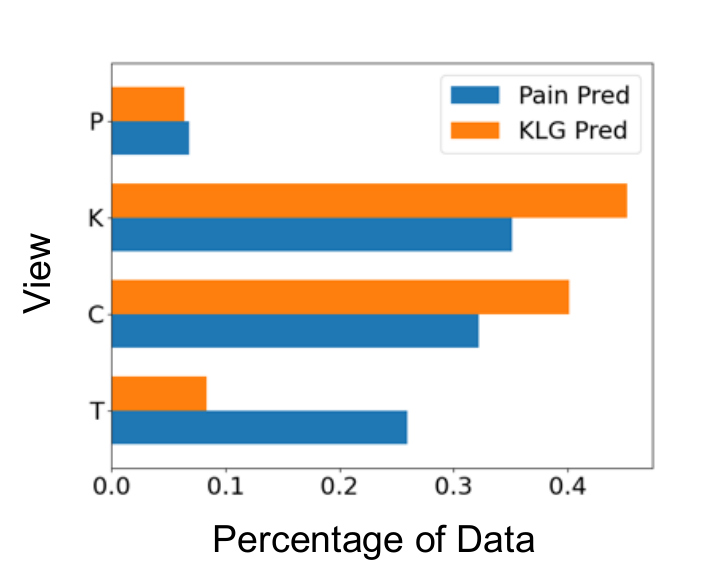}
      \label{fig: PercData}
  }
  \hfill
  \subfloat[]{
      \includegraphics[width=0.64\textwidth]{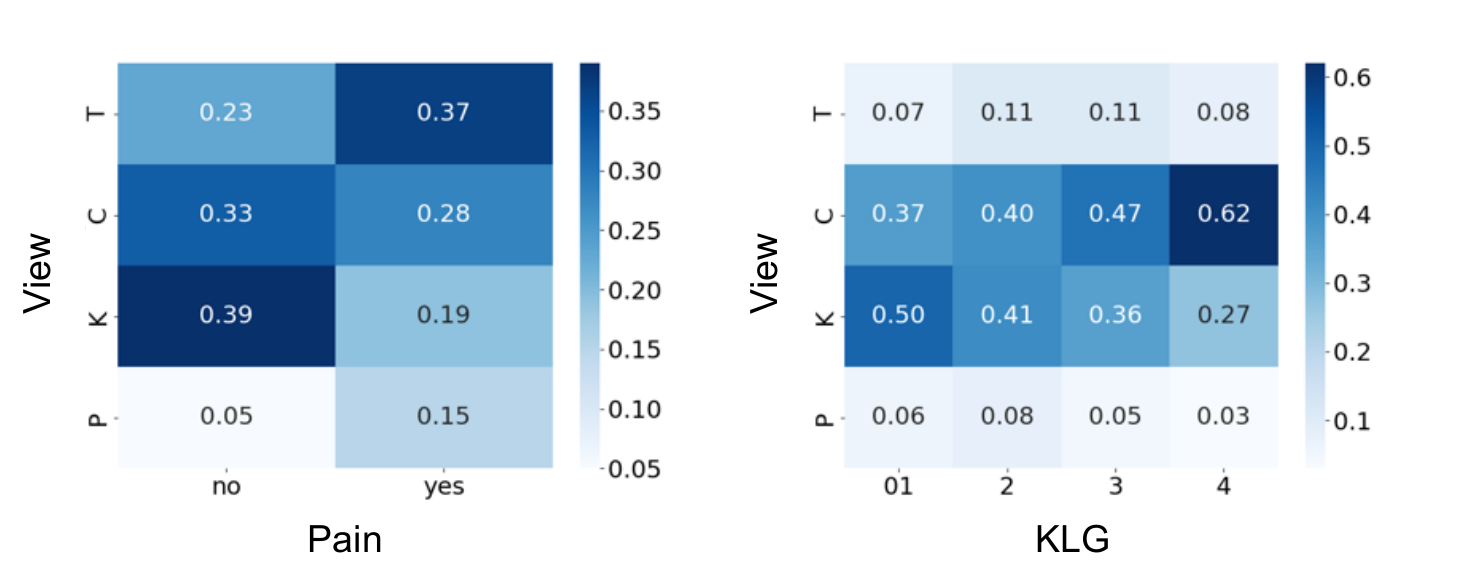}
      \label{fig: Heatmap}
  }
  \caption{Visualization of the most influential view for pain and KLG prediction. Left: Percentage of data where the view is deemed the most influential. Right: Normalized heatmaps showing the most influential view for each class.}
\end{figure}

\noindent\textit{How significant is each view in contributing to our unified model?}
Having a unified model allows us to easily determine the most pivotal view. We systematically exclude one at a time and observe the resulting change in prediction scores relative to the gold-standard labels. The view whose exclusion led to the worst performance is then considered the most crucial. Fig.~\ref{fig: PercData} shows that knee radiography emerged as the primary view for both tasks, followed by knee cartilage thickness maps. Tabular emerged as being helpful for pain prediction. % data are mostly self-reported, it might provide direct information of the patient's condition, which can be helpful for pain prediction.

\noindent\textit{Are there patterns between view importance and different prediction labels?}
Similar to assessing overall view significance, our unified model easily allows us to assess the view importance stratified by class. Fig.~\ref{fig: Heatmap} shows this view importance normalized by the number of instances in each class. We observe that knee radiography is the pivotal view in predicting non-pain instances, while tabular data emerged as more influential for instances associated with pain. For KLG prediction, cartilage thickness maps notably influenced higher grades, whereas knee images played a crucial role in lower grades. This observation aligns with the understanding that cartilage thinning becomes increasingly apparent with higher KLG, a characteristic readily discernible from cartilage thickness maps. Pelvis radiography emerged as the least impactful view across both experiments.

\section{Conclusion}
In this work, we proposed a unified model for longitudinal multi-modal multi-view prediction with missingness. Our model offers flexibility by accommodating various modalities, views, and timepoints, while adeptly handling missing data in the input. Through evaluation on the OAI dataset, we show the advantages of our unified model being able to generalize to different view combinations during evaluation. We also demonstrated the benefit of incorporating longitudinal data. Further, having a unified model allows us to easily probe and analyze the importance of different views for different prediction tasks. Future directions could include expanding the scope of our model by incorporating additional views. We also aim to implement an automatic view pruning technique, ensuring optimal prediction performance with the least number of views acquired.

\section{Acknowledgements}
This work was supported by NSF grants IIS2133595, DMS2324394, and NIH grants 1R01AR072013, 1R01AR082684, 1R01AA02687901A1, 1OT2OD032581-02-321. The work expresses the views of the authors, not of the NSF or NIH. The knee data were obtained from the controlled access datasets distributed from the Osteoarthritis Initiative (OAI), a data repository housed within the NIMH Data Archive. OAI is a collaborative informatics system created by NIMH and NIAMS to provide a worldwide resource for biomarker identification, scientific investigation and OA drug development. Dataset identifier: NIMH Data Archive Collection ID: 2343.

\bibliographystyle{splncs04}

\bibliography{main}

\begin{thebibliography}{10}
\providecommand{\url}[1]{\texttt{#1}}
\providecommand{\urlprefix}{URL }
\providecommand{\doi}[1]{https://doi.org/#1}

\bibitem{bhagwat2018modeling}
Bhagwat, N., Viviano, J.D., Voineskos, A.N., Chakravarty, M.M., Initiative, A.D.N., et~al.: Modeling and prediction of clinical symptom trajectories in {Alzheimer’s} disease using longitudinal data. PLoS computational biology  \textbf{14}(9),  e1006376 (2018)

\bibitem{cascarano2023machine}
Cascarano, A., Mur-Petit, J., Hernandez-Gonzalez, J., Camacho, M., de~Toro~Eadie, N., Gkontra, P., Chadeau-Hyam, M., Vitria, J., Lekadir, K.: Machine and deep learning for longitudinal biomedical data: a review of methods and applications. Artificial Intelligence Review  \textbf{56}(Suppl 2),  1711--1771 (2023)

\bibitem{mris}
Chen, B., Niethammer, M.: {MRIS}: A multi-modal retrieval approach for image synthesis on diverse modalities. In: MICCAI. pp. 271--281. Springer (2023)

\bibitem{chen2022recent}
Chen, X., Wang, X., Zhang, K., Fung, K.M., Thai, T.C., Moore, K., Mannel, R.S., Liu, H., Zheng, B., Qiu, Y.: Recent advances and clinical applications of deep learning in medical image analysis. MedIA  \textbf{79},  102444 (2022)

\bibitem{cui2019rnn}
Cui, R., Liu, M., Initiative, A.D.N., et~al.: {RNN}-based longitudinal analysis for diagnosis of {Alzheimer’s} disease. Computerized Medical Imaging and Graphics  \textbf{73},  1--10 (2019)

\bibitem{imagenet}
Deng, J., Dong, W., Socher, R., Li, L.J., Li, K., Fei-Fei, L.: {I}mage{n}et: A large-scale hierarchical image database. In: CVPR. pp. 248--255 (2009)

\bibitem{gibbons2010advances}
Gibbons, R.D., Hedeker, D., DuToit, S.: Advances in analysis of longitudinal data. Annual review of clinical psychology  \textbf{6},  79--107 (2010)

\bibitem{imagebind}
Girdhar, R., El-Nouby, A., Liu, Z., Singh, M., Alwala, K.V., Joulin, A., Misra, I.: {I}mage{b}ind: One embedding space to bind them all. In: CVPR. pp. 15180--15190 (2023)

\bibitem{resnet}
He, K., Zhang, X., Ren, S., Sun, J.: Deep residual learning for image recognition. In: CVPR. pp. 770--778 (2016)

\bibitem{dadp}
Huang, C., Xu, Z., Shen, Z., Luo, T., Li, T., Nissman, D., Nelson, A., Golightly, Y., Niethammer, M., Zhu, H.: {DADP}: dynamic abnormality detection and progression for longitudinal knee magnetic resonance images from the osteoarthritis initiative. MedIA  \textbf{77},  102343 (2022)

\bibitem{rotterdam}
Ikram, M.A., Brusselle, G., Ghanbari, M., Goedegebure, A., Ikram, M.K., Kavousi, M., Kieboom, B.C., Klaver, C.C., de~Knegt, R.J., Luik, A.I., et~al.: Objectives, design and main findings until 2020 from the {Rotterdam} study. European journal of epidemiology  \textbf{35},  483--517 (2020)

\bibitem{keefe2023patterns}
Keefe, T.H., Minnig, M.C., Arbeeva, L., Niethammer, M., Xu, Z., Shen, Z., Chen, B., Nissman, D.B., Golightly, Y.M., Marron, J., et~al.: Patterns of variation among baseline femoral and tibial cartilage thickness and clinical features: Data from the osteoarthritis initiative. Osteoarthritis and Cartilage Open  \textbf{5}(1),  100334 (2023)

\bibitem{klg}
Kellgren, J.H., Lawrence, J.: Radiological assessment of osteoarthrosis. Annals of the rheumatic diseases  \textbf{16}(4), ~494 (1957)

\bibitem{kohn2016classifications}
Kohn, M.D., Sassoon, A.A., Fernando, N.D.: Classifications in brief: {Kellgren-Lawrence} classification of osteoarthritis. Clinical Orthopaedics and Related Research  \textbf{474},  1886--1893 (2016)

\bibitem{hibehrt}
Li, Y., Mamouei, M., Salimi-Khorshidi, G., Rao, S., Hassaine, A., Canoy, D., Lukasiewicz, T., Rahimi, K.: {Hi-BEHRT}: Hierarchical transformer-based model for accurate prediction of clinical events using multimodal longitudinal electronic health records. IEEE journal of biomedical and health informatics  \textbf{27}(2),  1106--1117 (2022)

\bibitem{completer}
Lin, Y., Gou, Y., Liu, Z., Li, B., Lv, J., Peng, X.: {Completer}: Incomplete multi-view clustering via contrastive prediction. In: CVPR. pp. 11174--11183 (2021)

\bibitem{ukbiobank}
Littlejohns, T.J., Sudlow, C., Allen, N.E., Collins, R.: {UK} biobank: opportunities for cardiovascular research. European heart journal  \textbf{40}(14),  1158--1166 (2019)

\bibitem{dicnet}
Liu, C., Wen, J., Luo, X., Huang, C., Wu, Z., Xu, Y.: {DICNet}: Deep instance-level contrastive network for double incomplete multi-view multi-label classification. arXiv:2303.08358  (2023)

\bibitem{lmvcat}
Liu, C., Wen, J., Luo, X., Xu, Y.: Incomplete multi-view multi-label learning via label-guided masked view-and category-aware transformers. arXiv:2303.07180  (2023)

\bibitem{liu2022assessing}
Liu, Y., Yue, L., Xiao, S., Yang, W., Shen, D., Liu, M.: Assessing clinical progression from subjective cognitive decline to mild cognitive impairment with incomplete multi-modal neuroimages. MedIA  \textbf{75},  102266 (2022)

\bibitem{adamw}
Loshchilov, I., Hutter, F.: Decoupled weight decay regularization. arXiv:1711.05101  (2017)

\bibitem{marti2020survey}
Mart{\'\i}-Juan, G., Sanroma-Guell, G., Piella, G.: A survey on machine and statistical learning for longitudinal analysis of neuroimaging data in {Alzheimer’s} disease. Computer methods and programs in biomedicine  \textbf{189},  105348 (2020)

\bibitem{womac}
McConnell, S., Kolopack, P., Davis, A.M.: The {Western Ontario} and {McMaster} universities osteoarthritis index ({WOMAC}): a review of its utility and measurement properties. Arthritis Care \& Research: Official Journal of the American College of Rheumatology  \textbf{45}(5),  453--461 (2001)

\bibitem{clip}
Radford, A., Kim, J.W., Hallacy, C., Ramesh, A., Goh, G., Agarwal, S., Sastry, G., Askell, A., Mishkin, P., Clark, J., et~al.: Learning transferable visual models from natural language supervision. In: ICML. pp. 8748--8763 (2021)

\bibitem{gpt2}
Radford, A., Wu, J., Child, R., Luan, D., Amodei, D., Sutskever, I., et~al.: Language models are unsupervised multitask learners. OpenAI blog  \textbf{1}(8), ~9 (2019)

\bibitem{onecycle}
Smith, L.N., Topin, N.: Super-convergence: Very fast training of neural networks using large learning rates. In: Artificial intelligence and machine learning for multi-domain operations applications. vol. 11006, pp. 369--386 (2019)

\bibitem{saint}
Somepalli, G., Goldblum, M., Schwarzschild, A., Bruss, C.B., Goldstein, T.: {Saint}: Improved neural networks for tabular data via row attention and contrastive pre-training. arXiv:2106.01342  (2021)

\bibitem{kneel}
Tiulpin, A., Melekhov, I., Saarakkala, S.: {KNEEL}: Knee anatomical landmark localization using hourglass networks. In: ICCV Workshop. pp.~0--0 (2019)

\bibitem{transformer}
Vaswani, A., Shazeer, N., Parmar, N., Uszkoreit, J., Jones, L., Gomez, A.N., Kaiser, {\L}., Polosukhin, I.: Attention is all you need. NeurIPS  \textbf{30} (2017)

\bibitem{xu2022multi}
Xu, J., Tang, H., Ren, Y., Peng, L., Zhu, X., He, L.: Multi-level feature learning for contrastive multi-view clustering. In: CVPR. pp. 16051--16060 (2022)

\bibitem{zhou2023incomplete}
Zhou, Q., Zou, H., Jiang, H., Wang, Y.: Incomplete multimodal learning for visual acuity prediction after cataract surgery using masked self-attention. In: MICCAI. pp. 735--744 (2023)

\end{thebibliography}

\clearpage
\appendix

\section{OAI Data}  \label{sec: OAIOverview}
\begin{figure}
    \centering
    \includegraphics[width=.49\linewidth]{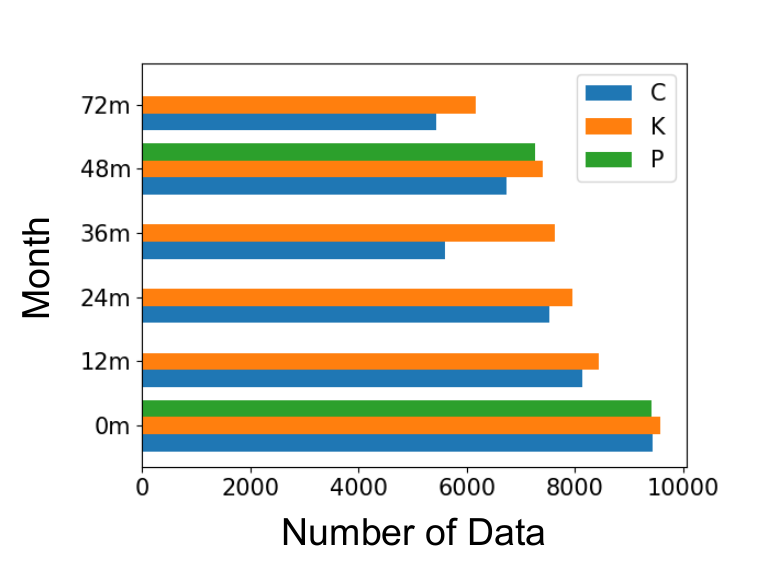}
    \caption{Data availability across modalities and views in the OAI dataset, up to $72m$. The $96m$ data points are excluded as they were not utilized as input. Pelvis data is only available at $0m$ and $48m$.}
    \label{fig: DataOverview}
\end{figure}

\section{Tabular Data}   \label{sec: Tabular}
\begin{table}[h]
    \begin{center}
    \resizebox{\textwidth}{!}{
    \begin{tabular}{|M{1.8cm}|M{4.8cm}||M{1.8cm}|M{4.8cm}|}
        \hline
        Attribute & Explanation & Attribute & Explanation \\
        \hhline{|====|}
        AGE & Age & SEX & Gender, male or female \\
        \hline
        RACE & Racial background & HISP & Hispanic or Latino \\
        \hline
        MARITST & Marital status & BMI & Body mass index \\
        \hline
        BPSYS & Blood pressure: systolic & BPDIAS & Blood pressure: diastolic  \\
        \hline
        EDCV & Highest grade or year of school completed & CEMPLOY & Current employment \\
        \hline
        CUREMP & Currently work for pay & INCOME2 & Yearly income ($>50K$ or $<50K$) \\
        \hline
        SMOKE & Have you smoked at least 100 cigarettes in entire life & DRNKAMT & How many alcoholic drinks in typical week, past 12 months \\
        \hline
        DRKMORE & Ever drink more beer, wine or liquor than do now & FAMHXKR & Mother, father, sister, or brother had knee repl surgery where all/part of knee replaced \\
        \hline
        MEDINS & Have any health insurance plan that pays for all or part of cost of prescription medicines & SIDE & Right or left side of the prediction\\
        \hline
    \end{tabular}
    }
    \end{center}
    \caption{Attributes and their explanations that were used for our tabular dataset. }
    \label{tabular_attr}
\end{table}

\clearpage

\section{Model Results}        \label{sec: PredResult}

\begin{table}[h]
    \begin{center}
    \resizebox{\textwidth}{!}{
    \begin{tabular}{|M{0.3cm}|M{0.3cm}|M{0.3cm}|M{0.3cm}|M{2.0cm}|M{2.0cm}|M{2.0cm}|M{2.0cm}|M{2.0cm}|M{2.0cm}|}
        \hline
        \multicolumn{4}{|c|}{Views} & \multicolumn{3}{c|}{Pain Prediction} & \multicolumn{3}{c|}{KLG Prediction} \\
        \hline
         T & C & K & P & AP & ROC  & Macro ACC & AP & ROC & Macro ACC \\
        \hline
        \cmark & & & & $0.302 \pm 0.003$ & $0.640 \pm 0.002$ & $0.600 \pm 0.003$ & $0.312 \pm 0.003$ & $0.618 \pm 0.005$ & $0.336 \pm 0.011$ \\ 
        \hline 
        & \cmark & & & $0.284 \pm 0.004$ & $0.661 \pm 0.003$ & $0.618 \pm 0.005$ & $0.660 \pm 0.005$ & $0.875 \pm 0.002$ & $0.642 \pm 0.007$ \\ 
        \hline 
        & & \cmark & & $0.325 \pm 0.004$ & $0.697 \pm 0.001$ & $0.648 \pm 0.005$ & $0.699 \pm 0.009$ & $0.881 \pm 0.003$ & $0.647 \pm 0.012$ \\ 
        \hline 
        \cmark & \cmark & & & $0.339 \pm 0.003$ & $0.696 \pm 0.002$ & $0.636 \pm 0.004$ & $0.672 \pm 0.005$ & $0.880 \pm 0.003$ & $0.652 \pm 0.008$ \\ 
        \hline 
        \cmark & & \cmark & & $0.357 \pm 0.003$ & $0.713 \pm 0.002$ & $\textbf{0.655} \pm \textbf{0.002}$ & $0.707 \pm 0.009$ & $0.886 \pm 0.002$ & $0.644 \pm 0.016$ \\ 
        \hline 
        \cmark & & & \cmark & $0.306\pm0.004$ & $0.653\pm0.003$ & $0.604\pm0.003$ & $0.315\pm0.003$ & $0.618\pm0.003$ & $0.319\pm0.008$ \\
        \hline
        & \cmark & \cmark & & $0.317 \pm 0.003$ & $0.696 \pm 0.002$ & $0.633 \pm 0.020$ & $0.702 \pm 0.004$ & $0.892 \pm 0.003$ & $0.662 \pm 0.014$ \\ 
        \hline 
        \cmark & \cmark & \cmark & & $0.352 \pm 0.003$ & $0.710 \pm 0.001$ & $0.642 \pm 0.008$ & $0.711 \pm 0.005$ & $0.896 \pm 0.002$ & $0.658 \pm 0.015$ \\ 
        \hline
        \cmark & \cmark & & \cmark & $0.343\pm0.004$ & $0.699\pm0.002$ & $0.631\pm0.007$ & $0.659\pm0.007$ & $0.875\pm0.003$ & $0.648\pm0.008$\\
        \hline
        \cmark & & \cmark & \cmark & $0.358\pm0.007$ & $0.712\pm0.004$ & $0.611\pm0.007$ &$0.693\pm0.010$ & $0.879\pm0.004$ & $0.646\pm0.009$ \\
        \hline
        & \cmark & \cmark &\cmark & $0.319\pm0.003$ & $0.695\pm0.004$ & $0.646\pm0.007$ & $0.696\pm0.010$ & $0.889\pm0.005$ & $0.643\pm0.011$ \\
        \hline
        \cmark & \cmark & \cmark & \cmark & $0.351\pm0.005$ & $0.710\pm0.002$ & $0.647\pm0.011$ & $0.694\pm0.004$ & $0.889\pm0.003$ & $0.657\pm0.009$ \\ 
        \hhline{|=|=|=|=|=|=|=|=|=|=|}
        \cmark & \cmark & \cmark & \cmark & $\textbf{0.358} \pm \textbf{0.003}$ & $\textbf{0.714} \pm \textbf{0.001}$ & $0.654 \pm 0.004$ & $\textbf{0.734} \pm \textbf{0.006}$ & $\textbf{0.903} \pm \textbf{0.001}$ & $\textbf{0.680} \pm \textbf{0.008}$ \\ 
        \hline
    \end{tabular}
    }
    \end{center}
    \caption{Mean $\pm$ STD results for pain and KLG prediction. The last row is our unified model, and all others are from the view-specific model. Our unified model not only allows for missing views but also boosts performance when using all available data. Incorporating more relevant views often leads to better predictions. Pelvis radiographs emerge as not very helpful. Further, cartilage thickness maps are not very helpful for pain prediction when knee radiographs are available.}%since labels are based on knee condition}
\end{table}

\section{Additional Unified and Specific Model Comparison}      
\label{sec:AdditionalPruning}

\begin{figure}[h!]
    \centering
    \includegraphics[width=0.85\textwidth]{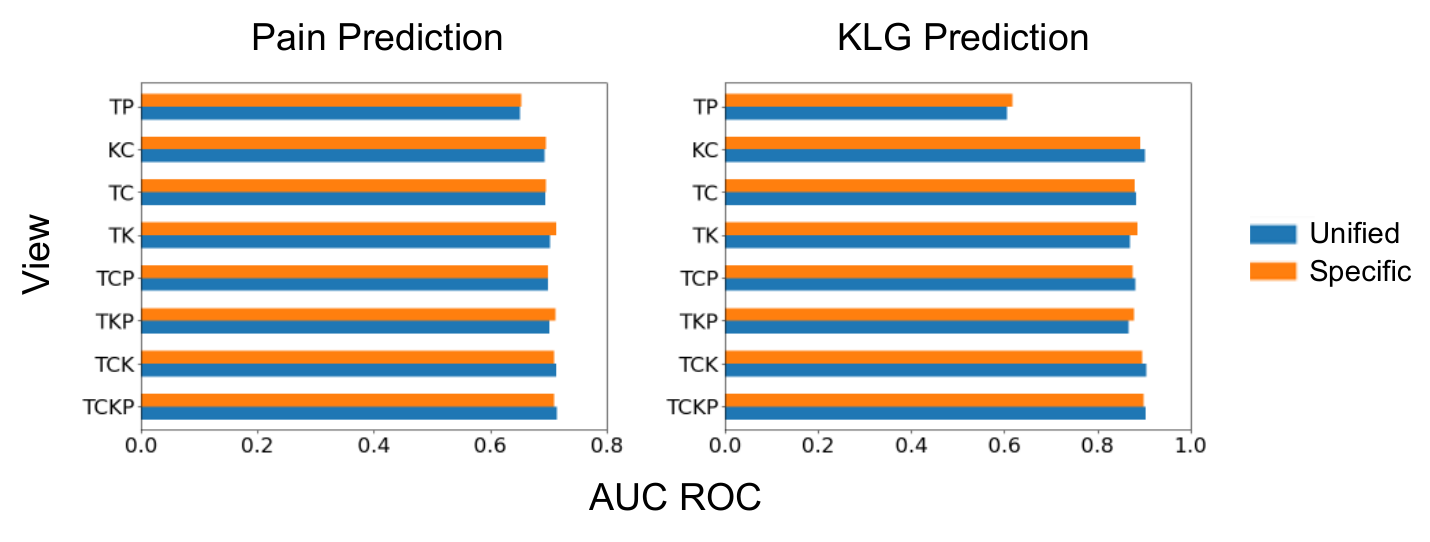}
    \caption{Comparison between our unified model and view-specific model on AUC ROC. Similar to AP, our unified model performs on par with view-specific models but is slightly better when all views are used.}
\end{figure}

% \end{document}

\end{document}